\documentclass[10pt,twocolumn,letterpaper]{article}

\usepackage{cvpr}              

\usepackage{graphicx}
\usepackage{amsmath}
\usepackage{amssymb}
\usepackage{booktabs}

\usepackage{nicefrac}       
\usepackage{microtype}      
\usepackage{slashbox}
\usepackage{bbm}
\usepackage{MnSymbol}
\usepackage{comment}
\usepackage{soul}
\usepackage{adjustbox}
\usepackage{balance}
\usepackage{array,multirow,graphicx}
\usepackage{subcaption}
\usepackage[table,xcdraw, x11names]{xcolor}
\usepackage[pagebackref,breaklinks,colorlinks]{hyperref}

\definecolor{1st}{HTML}{FFC71F}
\definecolor{2nd}{HTML}{FFE085}
\definecolor{3rd}{HTML}{FFF5D6}
\definecolor{bblue}{HTML}{4F81BD}
\definecolor{rred}{HTML}{C0504D}
\definecolor{ggreen}{HTML}{9BBB59}
\definecolor{ppurple}{HTML}{9F4C7C}
\definecolor{wwhite}{HTML}{FFFFFF}
\definecolor{pastelR}{HTML}{FBE5D6}
\definecolor{pastelG}{HTML}{E2F0D9}
\definecolor{pastelY}{HTML}{FFF2CC}
\definecolor{ggray}{gray}{0.9}
\definecolor{color_occ}{rgb}{0.0, 0.5, 0.0}
\definecolor{color_depth}{rgb}{0.0, 0.33, 0.73}

\AtBeginDocument{\hypersetup{citecolor=DodgerBlue3}}

\usepackage[capitalize]{cleveref}
\crefname{section}{Sec.}{Secs.}
\Crefname{section}{Section}{Sections}
\Crefname{table}{Table}{Tables}
\crefname{table}{Tab.}{Tabs.}

\def\cvprPaperID{656} 
\def\confName{CVPR}
\def\confYear{2022}

\begin{document}

\title{Instance-wise Occlusion and Depth Orders in Natural Scenes}
\author{
Hyunmin Lee\\
LG AI Research\\
{\tt\small hyunmin@lgresearch.ai}
\and
Jaesik Park\\
POSTECH GSAI \& CSE\\
{\tt\small jaesik.park@postech.ac.kr}
}
\maketitle

\begin{abstract}
In this paper, we introduce a new dataset, named \textbf{InstaOrder}, that can be used to understand the geometrical relationships of instances in an image. The dataset consists of 2.9M annotations of geometric orderings for class-labeled instances in 101K natural scenes. The scenes were annotated by 3,659 crowd-workers regarding (1) \textit{occlusion order} that identifies occluder/occludee and (2) \textit{depth order} that describes ordinal relations that consider relative distance from the camera. The dataset provides joint annotation of two kinds of orderings for the same instances, and we discover that the occlusion order and depth order are complementary. We also introduce a geometric order prediction network called \textbf{InstaOrderNet}, which is superior to state-of-the-art approaches. Moreover, we propose a dense depth prediction network called \textbf{InstaDepthNet} that uses auxiliary geometric order loss to boost the accuracy of the state-of-the-art depth prediction approach, MiDaS~\cite{Ranftl2020midas}. 
\end{abstract}

\section{Introduction}
\label{sec:introduction}
\begin{figure}[t]
\centering
\begin{subfigure}{.48\columnwidth}
  \centering
  \includegraphics[width=\columnwidth]{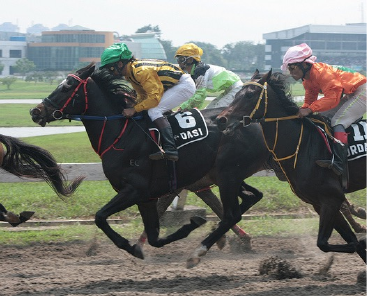}
  \caption{Image}
  \label{fig:overview_image}
\end{subfigure}
\begin{subfigure}{.48\columnwidth}
  \centering
  \includegraphics[width=\columnwidth]{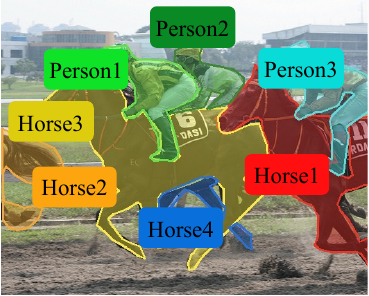}
  \caption{Instance masks}
  \label{fig:overview_mask}
\end{subfigure}
\begin{subfigure}{.48\columnwidth}
  \centering
  \includegraphics[width=\columnwidth]{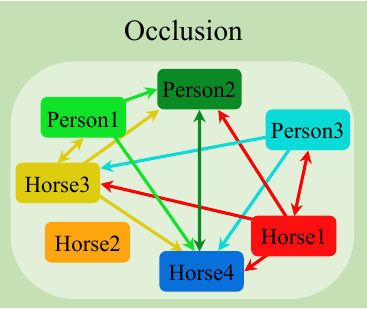}
  \caption{Occlusion order}
  \label{fig:overview_occ}
\end{subfigure}
\begin{subfigure}{.48\columnwidth}
  \centering
  \includegraphics[width=\columnwidth]{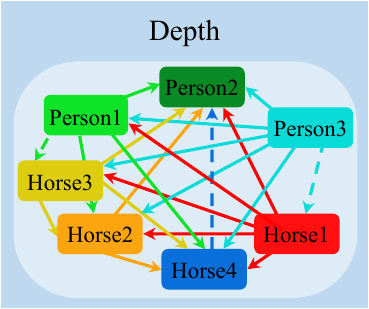}
  \caption{Depth order}
  \label{fig:overview_depth}
\end{subfigure}
\caption{Overview of the proposed \textsc{InstaOrder} dataset. (a and b) Example image of a cluttered scene and instance masks with class labels. (c) \emph{Occlusion order}. Arrows run from occluder to occludee. (d) \emph{Depth order}. Arrows point from close to far.}
\label{fig:i2d_overview}
\end{figure}

Understanding a scene from an image is a fundamental problem in computer vision. Deep learning-based approaches have achieved great success in various tasks, such as object detection~\cite{He2014det, Girshick2014det, Ren2015det, Girshick2015det, Dai2016det, Redmon2016det, Lin2017det, Beery2020det}, semantic segmentation~\cite{Noh2015eg, Liu2015seg, Long2015seg, Chen2016seg, Lin2016seg, Zhao2017seg, Yu2018seg, Voigtlaender2019seg, Liang2020seg, Vandenhende2020seg}, instance segmentation~\cite{Pinheiro2015iseg, Dai2016iseg, Dai2016iseg1, Oliveira2016iseg, Zhang2016iseg, Li2016iseg, He2017iseg, Liu2018iseg} and depth estimation~\cite{Eigen14depth, Chang18depth, Li19depth, Liu19depth, Godard19depth, Zhang19depth, Teed20depth, Johnston20depth, Xian20depth, Shu20depth}. More recently, approaches have inferred high-level information, such as amodal segmentation~\cite{Zhu17cocoa, Qi19kins, Yan19ovd, Zhou21ahp}, physics~\cite{Wu17physics}, and 3D-property recognition~\cite{Silberman12nyu, Xiao13sun3d, Song15sunrgbd, Hua16scenenn, Geiger2012driving, Cordts2016driving, Sun20driving, Yu20driving}. More importantly, many studies have emphasized the importance of understanding relationships between objects to learn high-level context~\cite{Rabinovich07relation, Parikh08relation, Galleguillos08relation, Jain10relation, Myeong12relation, Zoran15relation}. Given a natural image (Figure~\ref{fig:i2d_overview}a,~b), examples of such understanding would be `Horse3 occludes Person2.', `Horse1 and Person3 occlude each other.', or `Horse2 is closer than Person2.'.

For this purpose, we introduce a new large-scale dataset, called \textsc{InstaOrder}, for geometric scene understanding. The dataset has extensive annotations on \emph{geometric orderings between class-labeled instances in the natural scenes}. \textsc{InstaOrder} provides (1) \emph{Occlusion order} that determines objects that occlude others (occluders), and objects that are occluded (occludees), and (2) \emph{Depth order} that describes which object is closer or farther to the camera. \textsc{InstaOrder} is the first dataset that provides these two kinds of orders together from the same image.

\begin{figure*}[t]
\centering
\begin{subfigure}{0.64\columnwidth}
  \centering
  \includegraphics[height=1.3in]{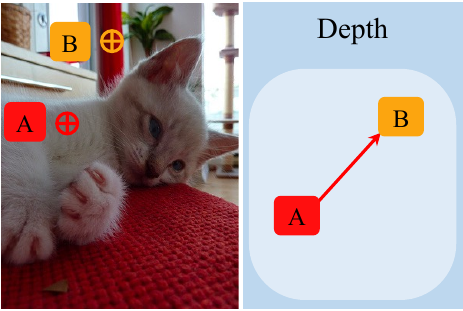}
  \caption{DIW~\cite{Chen16diw}}
  \label{fig:diw}
\end{subfigure}
\begin{subfigure}{.68\columnwidth}
  \centering
  \includegraphics[height=1.3in]{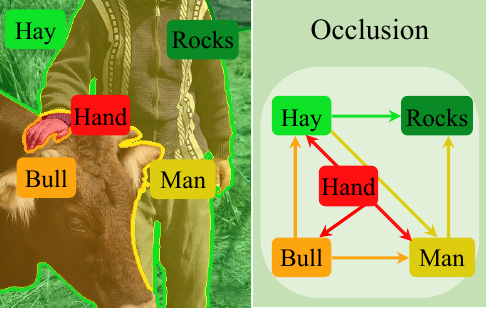}
  \caption{COCOA~\cite{Zhu17cocoa}}
  \label{fig:cocoa}
\end{subfigure}
\begin{subfigure}{0.68\columnwidth}
  \centering
  \includegraphics[height=1.3in]{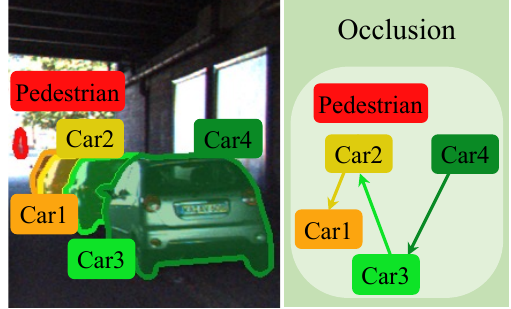}
  \caption{KINS~\cite{Qi19kins}}
  \label{fig:kins}
\end{subfigure}
\vspace{-2mm}
\caption{Overview of prior work. (a) DIW~\cite{Chen16diw} provides depth order of arbitrary points in a large-scale. (b) COCOA~\cite{Zhu17cocoa} and (c) KINS~\cite{Qi19kins} provide instance segmentation mask, instance class label, and instance-wise occlusion order.} 
\vspace{-2mm}
\label{fig:related} 
\end{figure*}

The two types of geometric relations can be expressed using directed graphs (Figure~\ref{fig:i2d_overview}c, d). Occlusion order and depth order are \emph{complementary} to each other, and \emph{neither alone can fully depict the geometric relationship in the cluttered scene.} For example, Horse2 in the occlusion graph (Figure~\ref{fig:i2d_overview}c) is isolated, so Horse2's depth is not clear without looking at the depth order graph (Figure~\ref{fig:i2d_overview}d). In contrast, looking only at the depth order graph does not demonstrate whether Horse1 occludes Horse3, whereas the occlusion order graph does provide such information.
Compared with other datasets shown in Figure~\ref{fig:related}, \textsc{InstaOrder} is the only large-scale and comprehensive dataset that provides instance segmentation mask, instance class label, occlusion, and depth order with delicate annotation of ordering types as shown as bidirectional edges and dashed edges.

\textsc{InstaOrder} is built on the COCO 2017~\cite{Lin14coco} dataset. A total of 3,659 crowd-workers annotated geometric ordering for 100,623 images having 503,939 instances, for a total of 2,859,919 depth and occlusion orderings. Such large-scale annotation distinguishes \textsc{InstaOrder} from the prior datasets that only cover occlusion order~\cite{Zhu17cocoa, Qi19kins} or depth order~\cite{Chen16diw}. In addition to its scale, \textsc{InstaOrder} introduces \emph{richer annotation on ambiguous cases} that had not been addressed before~\cite{Zhu17cocoa, Qi19kins, Chen16diw}. For example, \emph{bidirectional} order covers the case (Figure~\ref{fig:i2d_overview}) in which Horse3 and Person1 occlude each other. For depth order, in addition to $\{$closer, farther, or equal$\}$ orderings, we introduce \emph{distinct} and \emph{overlapping} depth orders. For example, some parts of Person1's left leg are closer than Horse3, whereas the right arm is farther (Figure~\ref{fig:i2d_overview}a). This case is annotated as overlap depth and displayed as a dashed line (Figure~\ref{fig:i2d_overview}d). The direction of the dashed line indicates that some part of Person1 (left leg) is nearer than any part of Horse3.

We also propose new networks called InstaOrderNet and InstaDepthNet. InstaOrderNet is used to recover instance-wise orders from an image. We show that InstaOrderNet achieves higher accuracy than state-of-the-art approaches, such as PCNet-M~\cite{Zhan20deocclusion} and OrderNet\textsuperscript{M+I}~\cite{Zhu17cocoa}. InstaDepthNet is used to predict a dense depth map from an image. With the proposed instance-wise disparity loss and the \textsc{InstaOrder} dataset, InstaDepthNet can boost the accuracy of MiDaS~\cite{Ranftl2020midas}, a state-of-the-art depth estimation network.

The contributions of this paper are:
\begin{itemize}
    \item We introduce the \textbf{\textsc{InstaOrder}} dataset that provides 2.9M of comprehensive instance-wise geometric orderings for 101K natural scenes. \textsc{InstaOrder} is the first dataset of both occlusion and depth order from the same image, with bidirectional occlusion order and delicate depth range annotations.
    \item We discover that occlusion and depth order are complementary, and that instance-wise orders are helpful for the monocular depth prediction task.
    \item We introduce \textbf{InstaOrderNet} for geometric order prediction and show its superior accuracy over state-of-the-arts. 
    In addition, we introduce \textbf{InstaDepthNet}, which demonstrates that the proposed auxiliary loss for geometric ordering can boost the depth prediction accuracy of the state-of-the-art approach, MiDaS~\cite{Ranftl2020midas}.
    \item The \textsc{InstaOrder} dataset, pre-trained model, and toolbox are available at \url{https://github.com/POSTECH-CVLab/InstaOrder}
\end{itemize}

\begin{table*}[t]
\setlength{\tabcolsep}{3pt}
\centering
\resizebox{2.05\columnwidth}{!}{
\small
\renewcommand*\arraystretch{1.1}
\begin{tabular}{lrccccccc}
\toprule
 & \# Img  & Scene & Source  & \# of classes  & Depth order & Occ. order & \# of annotations & Year \\ \hline\hline
DIW~\cite{Chen16diw} & \cellcolor{pastelG}495K & \cellcolor{pastelG}Natural Scenes& \cellcolor{pastelG} Flickr  & \cellcolor{pastelR}Not available & \cellcolor{pastelY}\begin{tabular}[c]{@{}c@{}}Two points per img\end{tabular}    & \cellcolor{pastelR}Not available  &    \cellcolor{pastelY}495K       & 2016               \\ 
COCOA~\cite{Zhu17cocoa}                & \cellcolor{pastelR}5K    & \cellcolor{pastelG}Natural Scenes& \cellcolor{pastelG}COCO~\cite{Lin14coco} & \cellcolor{pastelY}Not deterministic  & \cellcolor{pastelR}Not available                                                             & \cellcolor{pastelG}\begin{tabular}[c]{@{}c@{}}$\checkmark$ (instance)\end{tabular}  &  \cellcolor{pastelY}311K  & 2017           \\ 
KINS~\cite{Qi19kins}                 & \cellcolor{pastelR}15K  & \cellcolor{pastelY}Driving Scenes& \cellcolor{pastelG}KITTI~\cite{Geiger2012driving} & \cellcolor{pastelG}8    & \cellcolor{pastelR}Not available                                                                 & \cellcolor{pastelG}$\checkmark$ (instance)& \cellcolor{pastelG}{1.6M} & 2019           \\ \hline
\textsc{InstaOrder} & \cellcolor{pastelG}101K  & \cellcolor{pastelG}\textbf{Natural Scenes} & \cellcolor{pastelG}COCO~\cite{Lin14coco} & \cellcolor{pastelG}\textbf{80}   & \cellcolor{pastelG}\begin{tabular}[c]{@{}c@{}}\textbf{$\checkmark$ (instance)}\end{tabular} & \cellcolor{pastelG}\begin{tabular}[c]{@{}c@{}}\textbf{$\checkmark$ (instance)}\end{tabular}   & \cellcolor{pastelG}\textbf{2.9M}   &  Proposed \\ \bottomrule
\end{tabular}}
\vspace{-2mm}
\caption{Summary of related datasets available to the community. The colored cells indicate weak ($\mathrel{{\color{pastelR}\blacksquare}\!\!\!\!\square}$), moderate ($\mathrel{{\color{pastelY}\blacksquare}\!\!\!\!\square}$), or strong ($\mathrel{{\color{pastelG}\blacksquare}\!\!\!\!\square}$) points of each dataset. The proposed \textsc{InstaOrder} provides the largest amount of occlusion and depth annotations for various classes.}
\vspace{-2mm}

\label{table_related}
\end{table*}

\section{Related Work}
\label{sec:related}

\textbf{Datasets for occlusion orders.}
Understanding occlusion is proven to improve the ability of scene understanding in various computer vision tasks, such as  detection~\cite{Zhang18occ_det, Wang20occ_det}, instance segmentation~\cite{ke21occ, yuan21occ_seg}, depth estimation~\cite{Long20occ}, and optical flow estimation~\cite{Wang18occ_of,Hur19occ}. Recently, the concept of amodal perception has been emphasized, estimating the whole physical structure from a partial observation. Knowing occlusion order is crucial when inferring amodal masks. 

COCOA~\cite{Zhu17cocoa} is the first amodal dataset that contains both modal and amodal segmentation masks and their pair-wise occlusion orders. However, COCOA provides only 5,073 images, which is an insufficient number for data-driven approaches. Moreover, COCOA is designed with one-directional occlusion, and therefore splits instance masks if two instances occlude each other. In this procedure, annotators assign arbitrary labels to the new masks instead of assigning pre-defined class labels. KINS~\cite{Qi19kins} provides modal and amodal segmentation masks with relative occlusion order. KINS consists of 14,991 images, and it is built upon KITTI~\cite{Geiger2012driving} dataset. Therefore, all of its images are of driving scenes. 

\textsc{InstaOrder} is much larger than COCOA~\cite{Zhu17cocoa} and KINS~\cite{Qi19kins} (Table~\ref{table_related}). In addition, \textsc{InstaOrder} provides \emph{bidirectional occlusion}, whereas COCOA and KINS do not. We observed that bidirectional order facilitates the understanding of scenes.

\textbf{Occlusion order prediction.}
Tighe~\etal~\cite{Tighe14occlusion} build a histogram to predict occlusion overlap scores between two classes and solve quadratic integer programs. Zhu~\etal~\cite{Zhu17cocoa} proposed OrderNet\textsuperscript{M+I} that takes two masks and an image patch as input then produces pair-wise occlusion order in a supervised scheme. Zhan \etal~\cite{Zhan20deocclusion} proposed PCNet-M that recovers occlusion order in a self-supervised manner.

The proposed InstaOrderNet\textsuperscript{o} and InstaOrderNet\textsuperscript{o,d} can identify instances of ‘no occlusion’, ‘unidirectional occlusion’, and ‘bidirectional occlusion’. To the best of our knowledge, it is the first attempt in the field to identify three types simultaneously. We experimentally show that our InstaOrderNet\textsuperscript{o}, InstaOrderNet\textsuperscript{o,d} is more accurate than OrderNet\textsuperscript{M+I}~\cite{Zhu17cocoa} and PCNet-M~\cite{Zhan20deocclusion} networks in COCOA~\cite{Zhu17cocoa}, KINS~\cite{Qi19kins} and \textsc{InstaOrder} datasets.

\textbf{Datasets for depth maps.}
Advances in depth sensors have enabled extending 2D space to 3D space by collecting large-scale RGB-D datasets. For indoor environments, NYU Depth V2~\cite{Silberman12nyu}, SUN3D~\cite{Xiao13sun3d}, SUN RGB-D~\cite{Song15sunrgbd}, SceneNN~\cite{Hua16scenenn}, ScanNet~\cite{Hua16scenenn} datasets exist. For outdoor environments, KITTI~\cite{Geiger2012driving}, Cityscapes~\cite{Cordts2016driving}, Waymo Open Dataset~\cite{Sun20driving}, and BDD100K~\cite{Yu20driving} datasets exist. Although these datasets have enabled rapid progress in scene understanding, the scene types are restricted to either indoor or driving scenes. As a result, those datasets do not cover \emph{natural scenes}, in which diverse kinds of instances co-exist in an unconstrained setting. In addition, the depth value of transparent, specular, or distant objects is often not reliable because of the limitations of depth sensors. 

Recent datasets obtained geometric information for diverse scene types by crowd-sourcing~\cite{Chen16diw, Chen20oasis} or by applying photogrammetry methods~\cite{Li18megadepth, li19mannequin}. Stereo photos in the web~\cite{Xian18redweb, Xian20hr-wsi} or 3D movies~\cite{Ranftl2020midas} are used as another type of way to capture depth information. These datasets have been applied successfully to estimate a dense depth map from a single image, but they lack instance information and instance-wise relationships. In addition, the dataset for training a model is not publicly available~\cite{Ranftl2020midas}.

In contrast, \textsc{InstaOrder} covers instance masks, class labels, and instance-wise ordinal relationships. We experimentally show that instance-wise depth order improved modern monocular depth estimation networks like MiDaS~\cite{Ranftl2020midas}. 
\section{\textsc{InstaOrder} dataset}
\begin{figure*}[htb!]
    \centering
    \includegraphics[width=2\columnwidth]{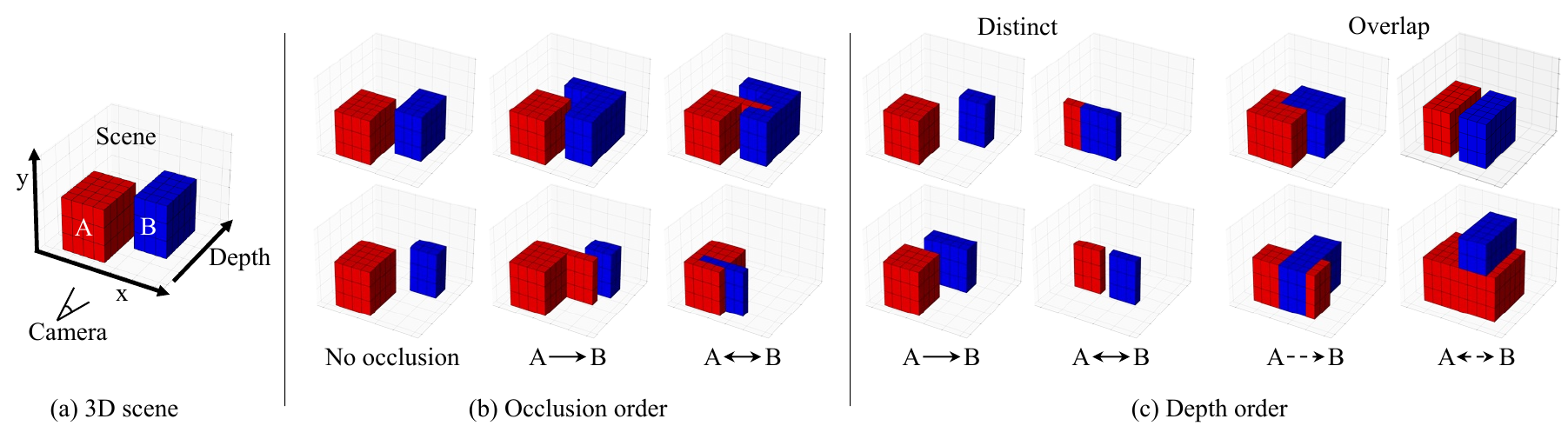}
    \vspace{-3mm}
    \caption{Ordering types defined for the \textsc{InstaOrder} dataset.
    (a) Camera looking at the scene of the two objects. 
    (b) Occlusion ordering is denoted considering the occluder and occludee.
    (c) Depth ordering. Distinct object pairs do not have overlapping depth regions, whereas overlapping pairs do. See Sec.~\ref{depth_order_labeling} for details.}
    \vspace{-3mm}
    \label{fig:orders_3d_view} 
\end{figure*}

\label{sec:method}
\subsection{Data Collection}
\textbf{Parent dataset.}
We annotate occlusion and depth orders upon COCO 2017~\cite{Lin14coco} dataset to get a benefit from large-scale instance labeling of natural scenes. Several other datasets also provide instance segmentation, such as LVIS~\cite{Gupta19lvis}, ADE20K~\cite{Zhou17ade20k}, Cityscapes~\cite{Cordts2016driving}. We decided to use COCO 2017 due to the following strengths: large-scale image set, covering diverse natural scenes, many instances in each image, and providing instance masks. We omit instances smaller than $25\times25$ pixels from the annotation because orders are often difficult to discern for tiny objects. We also discard inappropriate images for annotation, such as images with a single instance and collage images. As a result, images and instance masks in COCO 2017~\cite{Lin14coco} train set~(96,552 images) and validation set~(4,071 images) are used for the annotation.

\textbf{Annotation task.} \label{annotation_procedure}
As Todd~\etal~\cite{Todd03} stated, humans are good at judging relative depths. Inspired by this, our annotation procedure is designed as the task of requesting pairwise depth ordering between two instances in the same image. Both occlusion and depth annotation tasks start with guidelines, and then real examples appear as quizzes. Only annotators who passed all quizzes were allowed to participate in the annotation. Moreover, if a worker gives a wrong answer multiple times, the worker is dismissed. We provide a guideline to crowd-workers to annotate only the semantically meaningful instances (Sec. A3.2 in the supplement). 

\textbf{Minimizing dataset biases.}
We build our dataset with the following consideration to minimize dataset biases. 

\emph{(1) Class balance.} Our \textsc{InstaOrder} dataset reuses the images of the COCO 2017~\cite{Lin14coco} dataset, which was built using a careful image category decision and image collection mechanism to minimize dataset biases. The candidate image categories are gathered from frequently used words or PASCAL VOC~\cite{Mark10pascal}, and the decision is made by a vote on how one category is distinguishable from others. The COCO dataset is a collection of images from Flickr; to avoid collecting iconic photos, the process searched for images that had multiple keywords, such as 'dog + car’. 
\emph{(2) Crowdsourcing.} \textsc{InstaOrder} is collected using a sophisticated crowdsourcing engine. Thousands of people participated in the annotation of occlusion order or depth order, so the huge number of crowd-workers reduced the bias in the ordering annotations. To minimize diverged annotations, we asked two random workers to annotate every pairwise ordering. If the annotation results from two workers did not match, we invited additional workers until two of the workers made the same decision. We use \textbf{\emph{count}} to denote the number of participants per question, and our dataset provides count along with the occlusion and depth orderings to indicate the difficulty of the annotations.

\subsection{Ordering Types}
\label{depth_order_labeling}
Given a scene observed using a camera (Figure~\ref{fig:orders_3d_view}a), we identify occlusion and depth order. Occlusion order is determined by identifying the occluder and the occludee (Figure~\ref{fig:orders_3d_view}b). We utilize ‘no occlusion’ (no edge connection between A and B), ‘unidirectional occlusion’ (A occludes B: \textcolor{color_occ}{A$\rightarrow$B}; or B occludes A: \textcolor{color_occ}{B$\rightarrow$A}), and ‘bidirectional occlusion’ (A and B occlude each other: \textcolor{color_occ}{A$\longleftrightarrow$B}). 

Depth order denotes the relative distances of two objects from the camera. When an instance's depth range covers the other instance's depth range (Figure~\ref{fig:orders_3d_view}a), it is ambiguous to represent depth order with one of \{closer, farther, equal\}; thus, \emph{distinct, overlap} label is needed.
Depth order is annotated with a tuple of (x,y), where x$\in$\{closer, farther, equal\} and y$\in$\{distinct, overlap\}. Let’s denote instance $A$’s starting depth as $A_S$ and ending depth as $A_E$, where $A_S<=A_E$. \textcolor{color_depth}{A$\rightarrow$B} (distinct) is when $A_E<B_S$.
\textcolor{color_depth}{A$\longleftrightarrow$B} (distinct) is when $A_S=B_S=A_E=B_E$. A pair of instances in the same plane belong here (e.g. instances shown on TV). \textcolor{color_depth}{A$\dashrightarrow$B} (overlap) is when (i) $A_S < B_S<A_E$ or (ii) $A_S=B_S$ and $A_E<B_E$.
\textcolor{color_depth}{A$\dashleftarrow\dashrightarrow$B} (overlap) is when $A_S=B_S$, $A_E=B_E$ and $A_S\neq A_E$. Figure~\ref{fig:orders_3d_view}c shows such examples.

\begin{figure}[tbp]
    \centering
    \vspace{-2mm}
    \includegraphics[width=1.0\columnwidth]{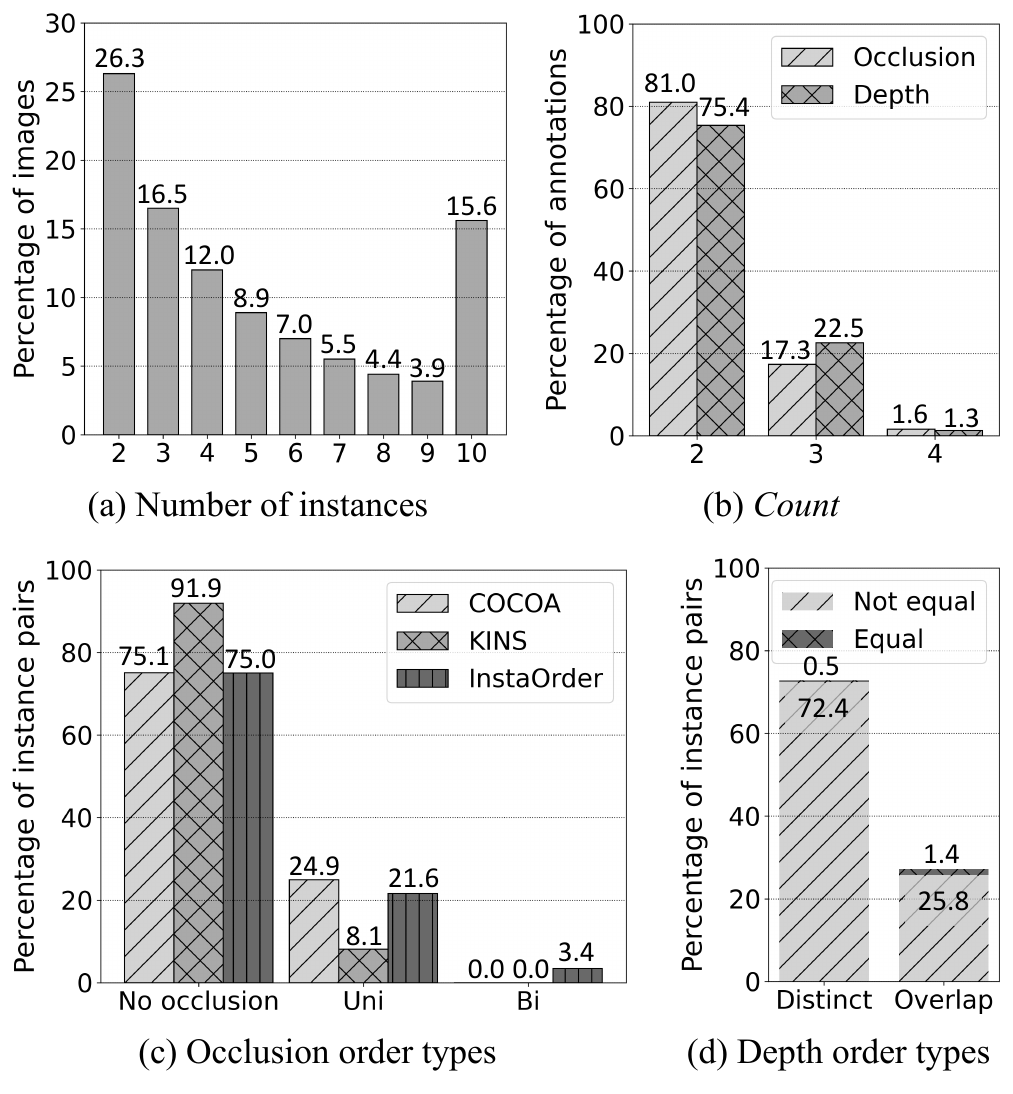}
    \vspace{-6mm}
    \caption{Statistics of the \textsc{InstaOrder} dataset. Sec.~\ref{statistics} for details.}
    \label{fig:statistic}
    \vspace{-4mm}
\end{figure}

\subsection{Statistics}
\label{statistics}
\textsc{InstaOrder} consists of 100,623 images with 503,939 instances that belong to 80 class categories, for a total of 2,859,919 instance-level occlusion and depth orders. Due to the limited annotation budget, we randomly selected ten instances if an image included more than ten instances. The histogram of instance number per image is shown in (Figure~\ref{fig:statistic}a).

The dataset was annotated by 1,549 crowd-workers for occlusion order, and by 2,110 for depth order. Of the instance pairs, 81\% were accepted by the first two annotators for occlusion ordering, whereas 75.4\% were accepted by the first two annotators for the depth ordering (Figure~\ref{fig:statistic}b). This comparison indicates that depth ordering was slightly harder to annotate than occlusion ordering.

\textsc{InstaOrder} has a similar distribution of occlusion order types to those of COCOA~\cite{Zhu17cocoa} or KINS~\cite{Qi19kins} (Figure~\ref{fig:statistic}c). In addition to unidirectional occlusion type, \textsc{InstaOrder} provides bidirectional occlusion order.
On depth order annotations, 72.9\% belonged to a distinct type and 27.2\% belonged to an overlap type (Figure~\ref{fig:statistic}d). The majority of depth orders belong to a ‘distinct’ and ‘not equal’ category (\textcolor{color_depth}{A$\rightarrow$B} or \textcolor{color_depth}{B$\rightarrow$A}), composing 72.4\% of total depth orders.

\subsection{Key Findings}
Here we provide interesting observations from the \textsc{InstaOrder}. These findings were observed in the comprehensive annotation of occlusion and depth order. Therefore, we highlight that these are new findings not discussed in previous literature~\cite{Chen16diw, Zhu17cocoa, Qi19kins}.

\begin{figure}[tbp]
    \centering
    \includegraphics[width=1.0\columnwidth]{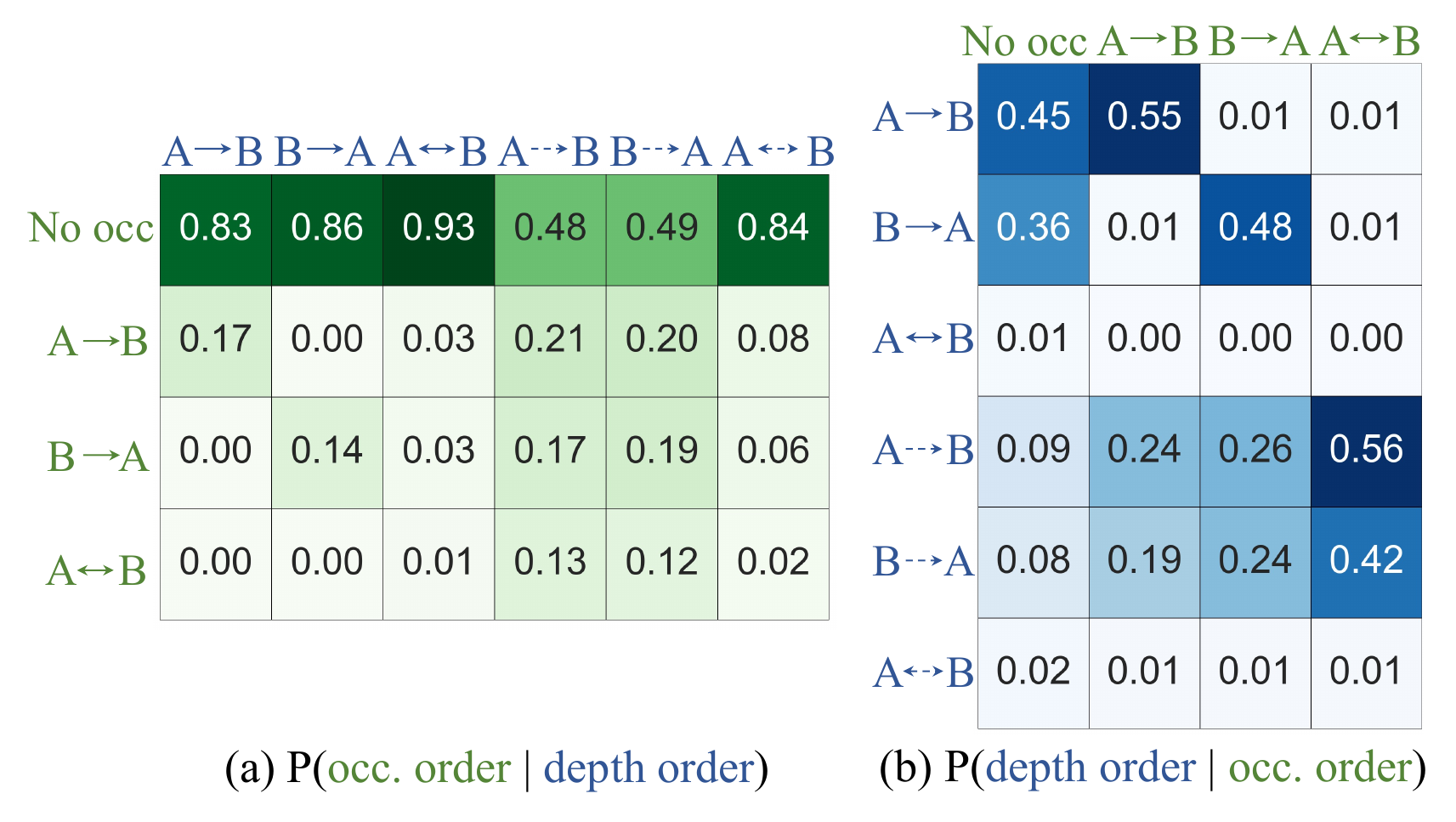}
    \vspace{-6mm}
    \caption{(a) Proportion of occlusion order given depth order and (b) vice versa. Each column is summed to one.}
    \vspace{-4mm}
    \label{fig:cond_prob}
\end{figure}

(1) \ul{\emph{Occlusion order and depth order should be annotated independently, because neither can indicate the other.}} 
We cannot perfectly infer depth order from occlusion order and vice versa. We demonstrate the claim with the correlation (Figure~\ref{fig:cond_prob}) between occlusion and depth order in the \textsc{InstaOrder} dataset. A pair of instances have occlusion and depth orders, so we can calculate P(occ. order $|$ depth order); the proportions of occlusion order types given depth order types (Figure~\ref{fig:cond_prob}a). For example, P(\textcolor{color_occ}{No occ} $|$ \textcolor{color_depth}{A$\rightarrow$B}) is 83\%. We can see that no “must happen correlation” occurs between two order types. For example, all types of occlusion order can occur when the depth order is \textcolor{color_depth}{A$\dashrightarrow$B}. Similar results are obtained using the proportions of depth order types given occlusion order type (Figure~\ref{fig:cond_prob}b). 
For reference, LabelMe~\cite{Russell08labelme} uses heuristics to infer occlusion order from instance masks. However, such heuristics are only applicable when masks intersect. In the \textsc{InstaOrder} dataset, only 16.4\% of mask pairs intersect, where we determine ‘intersect’ if two masks overlap more than ten pixels. Therefore, we cannot apply the heuristics to 83.6\% of non-intersecting instance pairs.

(2) \ul{\emph{Occlusion order and depth order are complementary to each other.}} We demonstrate this relationship in experiments (Tables~\ref{table_occ},~\ref{table_whdr}). A network trained with both occlusion and depth order is more accurate than baselines trained with individual orders. We think utilizing both types of orders eliminates cases that cannot happen. For example, (first column of Figure~\ref{fig:cond_prob}a) if the depth order is \textcolor{color_depth}{A$\rightarrow$B}, the occlusion orders \textcolor{color_occ}{B$\rightarrow$A} and \textcolor{color_occ}{A$\longleftrightarrow$B} cannot occur. Therefore, joint use of occlusion and depth orders provide rich supervision for comprehensive scene understanding.

(3) \ul{\emph{Bidirectional occlusion order is helpful.}} We conduct experiments with the \textsc{InstaOrder} dataset to verify the effect of bidirectional occlusion orders (Sec. A2.1 in the supplement). The result indicates methods that can determine bidirectional order distinguishes ambiguous cases better than methods that cannot determine bidirectional order.

\section{Methods}
This section introduces neural networks and a loss function that can be applied to the \textsc{InstaOrder} dataset. We present \textbf{InstaOrderNet}, which predicts instance-wise orders. Then we introduce a depth map prediction network \textbf{InstaDepthNet}, which gains accuracy with the proposed \textbf{instance-wise disparity loss}. The details of the network architecture are described in Sec. A4 in the supplement.

\subsection{Order Prediction}
\textbf{Occlusion order.}
The proposed InstaOrderNet\textsuperscript{o} takes pairwise instance masks and an image patch as input, then outputs occlusion order. InstaOrderNet\textsuperscript{o} is largely inspired by OrderNet\textsuperscript{M+I}~\cite{Zhu17cocoa}, so InstaOrderNet\textsuperscript{o} uses the pre-trained ResNet-50~\cite{He16resnet} backbone as OrderNet\textsuperscript{M+I} does.

OrderNet\textsuperscript{M+I}~\cite{Zhu17cocoa} classifies three types of occlusion order: \{No occlusion, A$\rightarrow$B, B$\rightarrow$A\}. The output dimension of OrderNet\textsuperscript{M+I} is [\texttt{batch\_size}, 3], trained with cross-entropy loss. 
On the other hand, the output dimension of InstaOrderNet\textsuperscript{o} is [\texttt{batch\_size}, 2] and it is trained with binary cross-entropy loss, $\mathcal{L}_{oo}$. InstaOrderNet\textsuperscript{o} solves two simple tasks: (1) ‘does A occludes B?’ and (2) ‘does B occludes A?’, answering two questions expresses four types of occlusion order: \{\textcolor{color_occ}{No occlusion, A$\rightarrow$B, B$\rightarrow$A, A$\longleftrightarrow$B}\}.

\textbf{Depth order.}
We introduce InstaOrderNet\textsuperscript{d} to predict depth order. InstaOrderNet\textsuperscript{d} takes pairwise instance masks and an image as input then produces depth orders. InstaOrderNet\textsuperscript{d} uses pre-trained ResNet-50 backbone, and it is trained with cross-entropy loss $\mathcal{L}_{do}$. The output dimension of the network is [\texttt{batch\_size}, 3], and three channels stand for \{\textcolor{color_depth}{A$\rightarrow$B, B$\rightarrow$A, A$\longleftrightarrow$B}\}.

\textbf{Occlusion and depth order.}
To demonstrate the effectiveness of jointly using occlusion and depth order, we introduce InstaOrderNet\textsuperscript{o,d}, which takes pairwise instance masks along with an image and produces \emph{both} occlusion and depth order. For a fair comparison, we build InstaOrderNet\textsuperscript{o,d} with the same neural architecture that was used in InstaOrderNet\textsuperscript{o} and InstaOrderNet\textsuperscript{d} except for the last fully-connected (FC) layer. Specifically, InstaOrderNet\textsuperscript{o,d} consists of two FC layers placed in parallel; one predicts occlusion order, and one predicts depth order (Figure A2 in the supplement). 

\subsection{Depth Map Prediction}\label{method_depth_map_prediction}
We also propose InstaDepthNet to show how instance-wise orderings can improve the state-of-the-art monocular depth estimation approach, MiDaS~\cite{Ranftl2020midas}. InstaDepthNet consists of two heads, one for instance-wise order prediction and the other for depth map prediction. The instance-wise order prediction head is composed of ResNet-50~\cite{He16resnet}. The depth map prediction head is composed of MiDaS-v2~\cite{Ranftl2020midas}, adopting pre-trained weights provided by authors. 

The InstaDepthNet architecture (Figure A3 in the supplement) has a modular design for the tasks. Therefore, the ordering prediction heads can be used for the training, and InstaDepthNet can produce a dense disparity map without instance masks during test time. We propose two versions of InstaDepthNet, such as InstaDepthNet\textsuperscript{d} and InstaDepthNet\textsuperscript{o,d} depending on the ordering types for the supervision.

We apply four loss functions to train InstaDepthNet. For the order prediction heads, we use binary cross-entropy loss $\mathcal{L}_{oo}$ for the occlusion order prediction head or cross-entropy loss $\mathcal{L}_{do}$ to the depth order prediction head.

For the depth map prediction head, we introduce \textbf{\emph{instance-wise disparity loss}} $\mathcal{L}_{disp}$. We denote depth order as $d_{AB}$ and set it to \{1, 0, -1\} when depth order is \{closer, equal, farther\}, respectively. Disparity is inversely proportional to depth, so when A is closer than B ($d_{AB}=1$), the disparity should be bigger for A than for B. $\mathcal{L}_{disp}$ penalizes violations of this relation by applying the proposed loss function:

\vspace{-4mm}
\begin{equation} \label{eq:loss_disp}
\begin{split}
  \mathcal{L}_{disp} \!=\! \frac{1}{2N} \!
  \sum_{i\in A \cup B} \! \Big\{
  \mathbbm{1}\Big(d_{AB} D'_{A}(i) \leq d_{AB} \max(D'_{B})\Big) + \\
  \mathbbm{1}\Big(d_{AB} D'_{B}(i) \geq d_{AB} \min (D'_{A})\Big)\Big\}, 
\end{split}
\end{equation}
where $i$ is a pixel in the area $A\cup B$, $D'_{A}$ is a predicted disparity map of A, $\mathbbm{1}(\cdot)$ is an indicator function, and $N$ is the number of pixels in $A\cup B$. We apply $\mathcal{L}_{disp}$ to distinct pairs because these orders are clear to supervise. 
We also use edge-aware smoothness loss~\cite{Godard17mono1}: 
$\mathcal{L}_{s} = \frac{1}{N} \sum_{i} \left | \partial_x D'(i)   \right | e^{-\left \| \partial_x I(i) \right \|} + \left | \partial_y D'(i) \right | e^{-\left \| \partial_y I(i) \right \|}$, where $I$ is an image, and $\partial_x$ and $\partial_y$ respectively are x- and y-directional image gradient operators.

The final loss is $\lambda_{0} \mathcal{L}_{oo} + \lambda_{1} \mathcal{L}_{do} + \lambda_{2} \mathcal{L}_{disp} + \lambda_{3} \mathcal{L}_{s}$. $\{\lambda_{0}, \lambda_{1}, \lambda_{2}, \lambda_{3}\}$ is set to \{0, 1, 1, 0.1\} for InstaDepthNet\textsuperscript{d}, and \{1, 1, 1, 0.1\} for InstaDepthNet\textsuperscript{o,d}. We conduct an ablation study on loss functions in Sec. A2.2 in the supplement.

\section{Experiments}
\subsection{Occlusion Order Recovery} 
\label{sec:occlusion_order_recovery}
\textbf{Baselines.} 
We compare the performance of the proposed InstaOrderNet\textsuperscript{o} with OrderNet\textsuperscript{M+I}~\cite{Zhu17cocoa} and PCNet-M~\cite{Zhan20deocclusion}. InstaOrderNet\textsuperscript{o} can process bidirectional occlusion order, whereas others cannot. Therefore we extend OrderNet\textsuperscript{M+I} to be able to predict bidirectional order and named it as OrderNet\textsuperscript{M+I}(ext.). For evaluating PCNet-M on COCOA~\cite{Zhu17cocoa} and KINS~\cite{Qi19kins} dataset, we use pre-trained weights provided by the authors. We use the official source code of PCNet-M for training and testing with \textsc{InstaOrder} dataset. We implement OrderNet\textsuperscript{M+I} from scratch, because the official code is not available\footnote{On the COCOA dataset, our OrderNet\textsuperscript{M+I} implementation achieves 89.1 recall, which is higher than the originally reported recall, 88.3.}.

\textbf{Simple approaches.} 
We also conduct experiments using a simple heuristic proposed by Zhu~\etal~\cite{Zhu17cocoa}. Specifically, given a pair of masks, the simple approach determines the occluder as the larger instance (the 'Area' method) or as the instance that is closer to the image bottom (the 'Y-axis' method). This experiment is intended to show that the simple prior cannot determine occlusion order well.

\textbf{Datasets.}
The experiments were conducted with three representative datasets that contain instance-wise occlusion order: COCOA~\cite{Zhu17cocoa}, KINS~\cite{Qi19kins}, and \textsc{InstaOrder}. For fairness, we train and test each method with the same dataset. For example, the second column in Table~\ref{table_occ} (top) indicates all methods are trained and tested with the KINS dataset.

\begin{table}[t]
\setlength{\tabcolsep}{3pt}
\centering
\resizebox{1\columnwidth}{!}{
\begin{tabular}{l | ccc | ccc}
\toprule
  & \multicolumn{3}{c|}{COCOA~\cite{Zhu17cocoa} dataset}   & \multicolumn{3}{c}{KINS~\cite{Qi19kins} dataset} \\ \cline{2-7}
Methods    & Recall $\uparrow$ & Prec. $\uparrow$ & F1 $\uparrow$    & Recall $\uparrow$ & Prec. $\uparrow$ & F1 $\uparrow$ \\ \hline\hline
PCNet-M~\cite{Zhan20deocclusion}    & {82.33} & {84.58}    & {82.80} &  {94.62} & {91.60}    & {92.59} \\
OrderNet\textsuperscript{M+I}~\cite{Zhu17cocoa}  & \textbf{89.12} & 83.91    & 85.63 &  98.33    & 93.45 & 95.19 \\  
\rowcolor{pastelY}
InstaOrderNet\textsuperscript{o}&  {88.60} & \textbf{85.38}    & \textbf{86.16} &  \textbf{98.70} &  \textbf{94.56}    &  \textbf{96.07} \\
\bottomrule
\end{tabular}}
\\
\resizebox{1\columnwidth}{!}{
\begin{tabular}{l|ccc|cc|ccc}
\toprule
\textsc{InstaOrder} dataset & \multicolumn{3}{c|}{Input} & \multicolumn{2}{c|}{Output} & \multicolumn{3}{c}{Occlusion acc. $\uparrow$} \\ \hline
Methods  &  \rotatebox[origin=c]{90}{Mask} & \rotatebox[origin=c]{90}{Image} & \rotatebox[origin=c]{90}{Category} & \rotatebox[origin=c]{90}{Occ. order} & \rotatebox[origin=c]{90}{Depth order} & Recall & Prec. & F1 \\\hline\hline
Area & $\checkmark$ &  &  & $\checkmark$ & & 56.33 & 71.55 & 59.67  \\ 
Y-axis & $\checkmark$ &  &  & $\checkmark$ & & 44.84 & 57.34 & 47.30 \\ \hline
PCNet-M~\cite{Zhan20deocclusion} & $\checkmark$ & $\checkmark$ & & $\checkmark$ & & 59.19 & 76.42 & 63.02 \\
OrderNet\textsuperscript{M+I}(ext.) & $\checkmark$ & $\checkmark$ & & $\checkmark$ & & 84.93 & 78.21 & 77.51 \\ \hline
InstaOrderNet\textsuperscript{o}(M) & $\checkmark$ &  &  & $\checkmark$ & &  87.35 &	79.07 &	78.98  \\ 
InstaOrderNet\textsuperscript{o}(MC) & $\checkmark$ &  & $\checkmark$ & $\checkmark$ & &  88.70&	78.21 & 79.18  \\ 
InstaOrderNet\textsuperscript{o}(MIC) & $\checkmark$ & $\checkmark$ & $\checkmark$ & $\checkmark$ & &  89.38 &	79.00 &	79.98  \\ 
\rowcolor{pastelY} InstaOrderNet\textsuperscript{o} & $\checkmark$ & $\checkmark$ & & $\checkmark$ & &  \textbf{89.39} & 79.83 & 80.65  \\ \hline
\rowcolor{pastelY}
InstaOrderNet\textsuperscript{o,d} & $\checkmark$ & $\checkmark$ & & $\checkmark$ & $\checkmark$ & 82.37 & \textbf{88.67} & \textbf{81.86} \\
\bottomrule
\end{tabular}}
\vspace{-2mm}
\caption{Occlusion order prediction results. We use COCOA~\cite{Zhu17cocoa} and KINS~\cite{Qi19kins} (top), and we use \textsc{InstaOrder} (bottom) for the experiments. We discuss methods highlighted in yellow in Sec.~\ref{sec:occlusion_order_recovery}.}
\vspace{-2mm}
\label{table_occ}
\end{table}

\textbf{Results.} 
To evaluate the occlusion order of every instance pair, we use \emph{Recall}, \emph{Precision} and \emph{F1} score (Table~\ref{table_occ}). In particular, we report the accuracy of the prediction of which of the two instances is an occluder, as done by OrderNet\textsuperscript{M+I}~\cite{Zhu17cocoa} and PCNet-M~\cite{Zhan20deocclusion}. PCNet-M is a network trained in a self-supervised manner, whereas OrderNet\textsuperscript{M+I} and our InstaOrderNet\textsuperscript{o} are trained in a supervised manner. Interestingly, PCNet-M\footnote{The recall reported in this paper is slightly different from the numbers appearing in PCNet-M because PCNet-M only consider neighboring instance mask pairs. In contrast, we used all pairs for the evaluation.} showed comparable accuracy to the supervised methods on both the COCOA~\cite{Zhu17cocoa} and KINS~\cite{Qi19kins} datasets (Table~\ref{table_occ}, top). 

\begin{figure*}[tb]
    \centering
    \includegraphics[width=\textwidth]{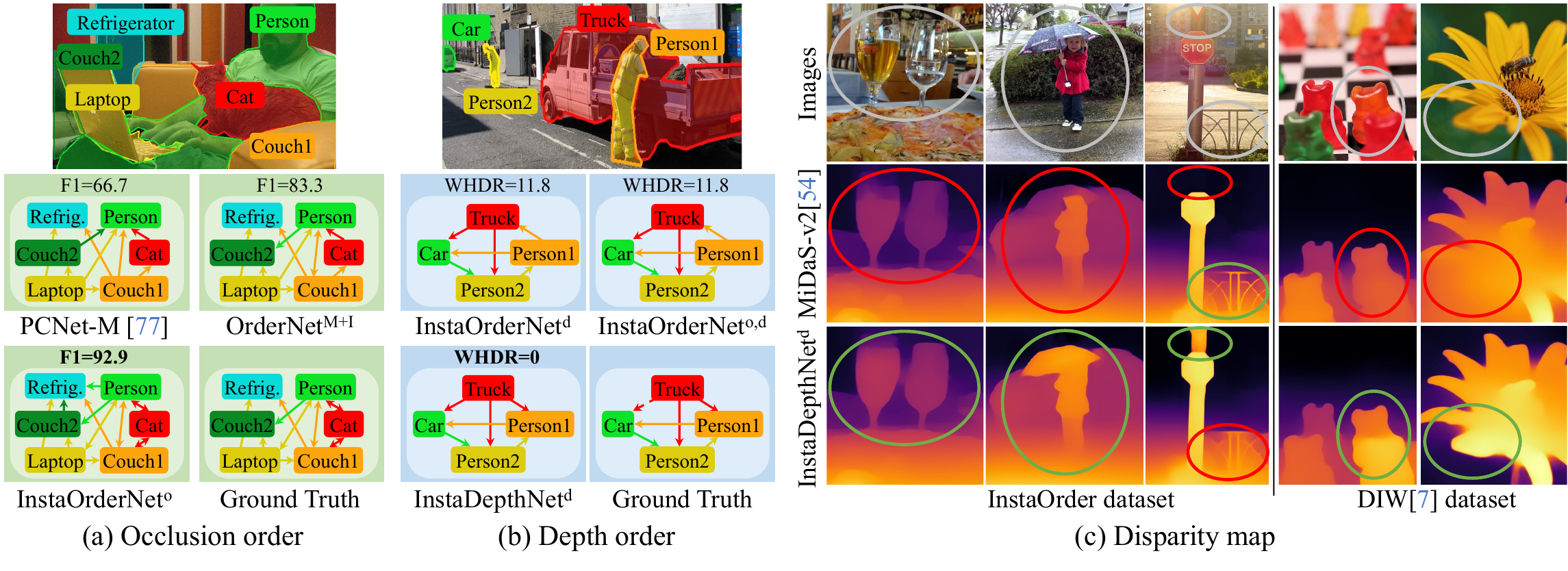}
    \vspace{-7mm}
    \caption{Qualitative results on the (a) occlusion, (b) depth order prediction. (c) Disparity maps generated by MiDaS-v2~\cite{Ranftl2020midas} and our InstaDepthNet\textsuperscript{d} on \textsc{InstaOrder} (left) and DIW~\cite{Chen16diw} (right). Red ellipses: unreasonable predictions; green ellipses: reasonable predictions.}
    \label{fig:qual_order_disp}
\end{figure*}

When the \textsc{InstaOrder} dataset was used (Table~\ref{table_occ}, bottom), all InstaOrderNet versions achieved significantly higher accuracy than the other methods. InstaOrderNet\textsuperscript{o} is a simple extension of OrderNet\textsuperscript{M+I}, but achieves higher accuracy because it converts multi-class classification to the multi-label classification problem. InstaOrderNet\textsuperscript{o,d} is more accurate than InstaOrderNet\textsuperscript{d}. We can infer that occlusion and depth order are not independent, but provide \emph{complementary} information for scene understanding. We observed that the accuracy of PCNet-M depends on instance mask quality and thus resulted in a low accuracy on the \textsc{InstaOrder} dataset. 
We show qualitative results on occlusion order prediction (Figure~\ref{fig:qual_order_disp}a), and InstaOrderNet\textsuperscript{o} showed the best accuracy. Even though OrderNet\textsuperscript{M+I}(ext.) is an extended network to predict bidirectional occlusion order, but still missed most bidirectional occlusion orders.

\textbf{Various input/output configurations.} We conduct experiments with different types of inputs, such as image and category labels (Table~\ref{table_occ}, bottom). When we provide a category label to the mask, we assign appropriate category IDs to the masked areas. For the occlusion order prediction task, a network that uses the mask as a sole input (InstaOrderNet\textsuperscript{o}(M)) is even comparable to the network that uses both mask and image (InstaOrderNet\textsuperscript{o}); a similar result was reported by Zhu~\etal~\cite{Zhu17cocoa}. We speculate that the mask provides enough clues to determine occlusion order. Moreover, we conduct an ablation study on bidirectional occlusion order (Sec.~A2.1 in the supplement).

\subsection{Depth Order Recovery}\label{depth_order_recovery}
\textbf{Baselines.}
To the best of our knowledge, no existing method directly analyzes depth ordering for instances. 
Therefore, we propose two baselines that use a state-of-the-art depth map estimation network, MiDaS-v2\cite{Ranftl2020midas}, for the comparison of depth order prediction accuracy. The baseline approach, MiDaS(Mean), uses the instance-wise mean of the disparity predicted by MiDaS-v2. Similarly, MiDaS(Median) uses the instance-wise median value. This choice is guided by an assumption that instance-wise mean or median values represent instance-wise distance\footnote{Instance segmentation masks of COCO 2017 are not perfect, so for reliability of comparison, we ignore the top and bottom 5\% disparity values.}. We compare baselines with the proposed InstaOrderNet\textsuperscript{d} (Table~\ref{table_whdr}). We also evaluate simple approaches (Area, Y-axis). 

\textbf{WHDR.} We evaluate the results using Weighted Human Disagreement Rate (WHDR)~\cite{bell14whdr}, which represents the percentage of weighted disagreement between ground truth $d$ and predicted depth order $d'$. The weights are proportional to the confidence of each annotation. Here, we use the inverse of \emph{count} multiplied by the minimum number of participants. We evaluate WHDR on each of \{distinct, overlap, all\} categories separately; which is defined as follows:
WHDR = $\frac{\sum_{AB} w_{AB}\cdot \mathbbm{1}(d'_{AB} \neq d_{AB})}{\sum_{AB} w_{AB}}$, where $w_{AB} = \frac{2}{\emph{count}_{AB}}$. 

\begin{table}[tb]
\setlength{\tabcolsep}{3pt}
\begin{center}
\resizebox{\columnwidth}{!}{
\renewcommand*\arraystretch{1.1}
\begin{tabular}{l|ccc|ccc|ccc}
\toprule
     & \multicolumn{3}{c|}{Input}& \multicolumn{3}{c|}{Output} & \multicolumn{3}{c}{WHDR $\downarrow$} \\ \cline{2-10}
Methods        & \rotatebox[origin=c]{90}{Mask} & \rotatebox[origin=c]{90}{Image} & \rotatebox[origin=c]{90}{Category} & \rotatebox[origin=c]{90}{Occ. order} & \rotatebox[origin=c]{90}{Depth order} & \rotatebox[origin=c]{90}{Disp. map} & \rotatebox[origin=c]{90}{Distinct} & \rotatebox[origin=c]{90}{Overlap} & \rotatebox[origin=c]{90}{All} \\ \hline\hline
Area & $\checkmark$ & & & & $\checkmark$ &  & \multicolumn{1}{r}{30.90} & 35.66 & 32.19\\
Y-axis & $\checkmark$ & & & & $\checkmark$ &  & \multicolumn{1}{r}{22.19} & 39.04 & 29.20 \\ \hline
MiDaS(Mean)~\cite{Ranftl2020midas} & & $\checkmark$ &   &  &   & $\checkmark$ & \multicolumn{1}{r}{10.42} & 37.67 & 21.70 \\
MiDaS(Median)~\cite{Ranftl2020midas} & & $\checkmark$ &  &  &   & $\checkmark$ & \multicolumn{1}{r}{10.31} & 36.08 & 20.92 \\
\hline
InstaOrderNet\textsuperscript{d}(M) & $\checkmark$ & &  & & $\checkmark$ & &  \multicolumn{1}{r}{22.96} & 30.46 & 25.23  \\
InstaOrderNet\textsuperscript{d}(MC) & $\checkmark$ &  &$\checkmark$  & & $\checkmark$ & & \multicolumn{1}{r}{23.19} & 28.56 & 36.45  \\ 
InstaOrderNet\textsuperscript{d}(MIC) & $\checkmark$ & $\checkmark$ & $\checkmark$ & & $\checkmark$ & & \multicolumn{1}{r}{13.33} & 26.60 & 17.89 \\ 
\rowcolor{pastelY}
InstaOrderNet\textsuperscript{d}     & $\checkmark$     & $\checkmark$ &  & & $\checkmark$ & &  \multicolumn{1}{r}{12.95} & 25.96 & 17.51  \\ \hline
\rowcolor{pastelY}
InstaOrderNet\textsuperscript{o,d} & $\checkmark$ & $\checkmark$ & & $\checkmark$ & $\checkmark$ & & 11.51 & 25.22 & 15.99 \\ \hline
InstaDepthNet$^{\text{d}}$(Mean) & $\checkmark$ & $\checkmark$  & & & $\checkmark$ & $\checkmark$ &  \multicolumn{1}{r}{9.80} & 37.97 & 21.46 \\
InstaDepthNet$^{\text{d}}$(Median) & $\checkmark$ & $\checkmark$  & & & $\checkmark$ & $\checkmark$ & \multicolumn{1}{r}{9.29} & 36.07 & 20.41 \\
\rowcolor{pastelY}
InstaDepthNet\textsuperscript{d}   & $\checkmark$ & $\checkmark$  & & & $\checkmark$ & $\checkmark$ &  \multicolumn{1}{r}{7.25} & 23.34 & 12.94 \\ \hline
\rowcolor{pastelY}
InstaDepthNet\textsuperscript{o,d}   & $\checkmark$ & $\checkmark$ & & $\checkmark$ &$\checkmark$ & $\checkmark$ &  \multicolumn{1}{r}{\textbf{7.00}} & \textbf{23.29} & \textbf{12.72} \\
\bottomrule   
\end{tabular}}
\end{center}
\vspace{-4mm} 
\caption{Depth order prediction results. We train and test networks with various input and output configurations. We discuss methods highlighted in yellow in Sec.~\ref{depth_order_recovery} and Sec.~\ref{depth_map_prediction}.}
\vspace{-2mm} 
\label{table_whdr}
\end{table}

\textbf{Results.} 
Both MiDaS(Mean, Median) achieved notable WHDR (Table~\ref{table_whdr}) even though \textsc{InstaOrder} is an unseen dataset to MiDaS~\cite{Ranftl2020midas}. Although trained with the \textsc{InstaOrder} dataset, InstaOrderNet\textsuperscript{d} had inferior (higher) WHDR than MiDaS(Mean, Median) for the distinct instances. We observe that instance-wise depth ordering must involve the disparity map prediction task, as InstaDepthNet does. InstaDepthNet is trained for dense disparity map prediction as well as order predictions. We believe that such comprehensive tasks give plentiful supervision for distinct instances and bring significant accuracy gain. As observed in occlusion order recovery results, InstaOrderNet\textsuperscript{o,d} is superior to InstaOrderNet\textsuperscript{d}; this indicates occlusion and depth order are complementary information.

When the image is not used as an input, the accuracy degrades (InstaOrderNet\textsuperscript{d}(M, MC)). The results are different from the previous experiment with occlusion order. We think the image gives a global context to determine relative depth order correctly. Qualitative results on depth order prediction (Figure~\ref{fig:qual_order_disp}b) indicate InstaDepthNet\textsuperscript{d} is better at figuring out tricky relations, such as \textcolor{color_depth}{Truck$\dashedrightarrow$Person}.

\subsection{Depth Map Prediction}
\label{depth_map_prediction}
\textbf{Testing with InstaOrder dataset.}
We further demonstrate that occlusion and depth orders can be used to increase the accuracy of a depth estimation network. We compare the disparity map predicted by MiDaS~\cite{Ranftl2020midas} and InstaDepthNet\textsuperscript{d} using (Sec.~\ref{depth_order_recovery} {Baselines}) the mean and median scheme. (Table~\ref{table_whdr}) InstaDepthNet$^{\text{d}}$(Mean, Median), were both more accurate than MiDaS(Mean, Median). 

We compare the qualitative result of the disparity map estimated by MiDaS-v2 with InstaDepthNet\textsuperscript{d} (Figure~\ref{fig:qual_order_disp}c). Human annotation in \textsc{InstaOrder} for challenging objects (transparent glasses) and instance-wise depth order helps to correct wrong disparity prediction. 

\textbf{Testing with unseen datasets.} 
First, we compare the depth order accuracy on the DIW~\cite{Chen16diw} dataset (Table~\ref{table_unseen}, top). We compare  InstaDepthNet\textsuperscript{d}, which was trained on the \textsc{InsataOrder} dataset, whereas MiDaS-v2 was trained using numerous 3D movies. InstaDepthNet\textsuperscript{d} showed superior disparity maps (Figure~\ref{fig:qual_order_disp}c, right); this result supports the value of the proposed \textsc{InstaOrder} dataset and the instance-wise disparity loss (Eq.~\ref{eq:loss_disp}). 

\begin{table}[t]
\setlength{\tabcolsep}{5pt}
\begin{center}
\resizebox{\columnwidth}{!}{
\begin{tabular}{l|ccc|ccc|ccc}
\toprule
DIW~\cite{Chen16diw} & \multicolumn{3}{c|}{Input}& \multicolumn{3}{c|}{Output} & \multicolumn{3}{c}{Accuracy} \\ \hline
Methods        & \rotatebox[origin=c]{90}{Mask} & \rotatebox[origin=c]{90}{Image} & \rotatebox[origin=c]{90}{Category} & \rotatebox[origin=c]{90}{Occ. order} & \rotatebox[origin=c]{90}{Depth order} & \rotatebox[origin=c]{90}{Disp. map} & \rotatebox[origin=c]{90}{\# Correct $\uparrow$} & \rotatebox[origin=c]{90}{\# Wrong $\downarrow$} & \rotatebox[origin=c]{90}{WHDR $\downarrow$} \\ \hline\hline
MiDaS-v2~\cite{Ranftl2020midas}  & & $\checkmark$  & & &  & $\checkmark$ &  64,723      & 9,718     & 13.06    \\
\rowcolor{pastelY}
InstaDepthNet\textsuperscript{d}   & & $\checkmark$ & & & &$\checkmark$  & \textbf{65,317}      & \textbf{9,124}     & \textbf{12.26}\\
\bottomrule
\end{tabular}}
\end{center}
\vspace{-3.5mm}
\setlength{\tabcolsep}{1pt}
\resizebox{1\columnwidth}{!}{
\begin{tabular}{l|ccc|ccc}
\toprule
KITTI~\cite{Geiger2012driving} & \multicolumn{3}{c|}{Error $\downarrow$}   & \multicolumn{3}{c}{Accuracy $\uparrow$} \\ \hline
Methods  & Abs Rel & Sq Rel & RMSE log & $\delta<$1.25 & $\delta<$1.25\textsuperscript{2} & $\delta<$1.25\textsuperscript{3} \\ \hline\hline
MiDaS-v2~\cite{Ranftl2020midas}    & 0.16 & 1.47 & \textbf{0.20} &\textbf{ 0.81} & 0.95 & 0.98 \\
\rowcolor{pastelY}
InstaDepthNet\textsuperscript{d} & \textbf{0.15} & \textbf{1.27} & \textbf{0.20} & 0.80 & \textbf{0.95} & \textbf{0.99} \\ \bottomrule
\end{tabular}}
\caption{Evaluation of predicted disparity maps using unseen datasets (top table: DIW~\cite{Chen16diw}, bottom table: KITTI~\cite{Geiger2012driving}).}
\vspace{-2mm}
\label{table_unseen}      
\end{table}

We also test two approaches with the KITTI dataset~\cite{Geiger2012driving} (Table~\ref{table_unseen}, bottom). The predicted disparity maps of both approaches are not on a metric scale, so we adopt per-image median ground truth scaling~\cite{Godard19depth}, and report statistical metrics~\cite{Eigen14depth} (details in Sec.~A1 in the supplement). Overall, InstaDepthNet\textsuperscript{d} was slightly better than MiDaS.

\textbf{Limitation.} 
InstaDepthNet\textsuperscript{d} did yield some problematic results (Figure~\ref{fig:qual_order_disp}c, red ellipse in the middle example). The confusion occurs because instance masks in our parent dataset, COCO 2017~\cite{Lin14coco} do not fully segment objects that have holes. We leave this problem to future work. 

\section{Discussion}
\label{sec:discussion} 
We introduce \textsc{InstaOrder} dataset and propose various order prediction networks. Our dataset has several benefits compared to DIW~\cite{Chen16diw}, COCOA~\cite{Zhu17cocoa}, or KINS~\cite{Qi19kins} in terms of scale, classes, and order types. We demonstrate the effectiveness of jointly using occlusion order and depth order. We show that the state-of-the-art depth map prediction approach can be improved by using the proposed auxiliary loss for instance-wise ordering. 

We plan to study the benefit of \textsc{InstaOrder} for the tasks beyond depth estimation. For example, as panoptic segmentation studies~\cite{Liu19panoptic, Lazarow20panoptic, Chen20panoptic} gain accuracy by figuring out the occlusion order between objects, \textsc{InstaOrder} can benefit the task by explicitly reasoning the occlusion order. In addition, \textsc{InstaOrder} can help image captioning or VQA tasks. Specifically, (Figure\ref{fig:i2d_overview}c,d) "Horse1" and "Person3" are occluding each other, then we can infer they are interacting. Moreover, we can make a question and answer like "Who is behind Person 1?". Image generation studies~\cite{Johnson18imagegen, Ashual19imagegen} use scene graphs for generating images. It would be interesting to create images by considering occlusion, depth order, and scene graph. Moreover, as Zhan~\etal~\cite{Zhan20deocclusion} manipulate images by controlling occlusion order, \textsc{InstaOrder} can also be used for image manipulation.

\vspace{2mm}
\textbf{Acknowledgments.} We thank Qualcomm for the generous support and SELECTSTAR for the data collection help. This work was supported by IITP grants 2021-0-00537 (Visual common sense) and 2019-0-01906 (Artifcial Intelligence Graduate School Program (POSTECH)) funded by the Korea government (MSIT).

\newpage
\renewcommand{\theequation}{A\arabic{equation}}
\renewcommand{\thetable}{A\arabic{table}}
\renewcommand{\thefigure}{A\arabic{figure}}
\renewcommand{\thesection}{A\arabic{section}}
\renewcommand{\thesubsection}{A\arabic{section}.\arabic{subsection}}

\section*{\Large{Appendices}}
\setcounter{section}{0}
\setcounter{figure}{0}    
\setcounter{table}{0}    
\setcounter{equation}{0}    

\section{Evaluation Metrics}
\subsection{Occlusion order recovery}
We evaluate the occlusion order of every instance pair using recall, precision, and F1 score. In particular, we report the accuracy of predicting which of the two instances is an occluder, as done in OrderNet\textsuperscript{M+I}~\cite{Zhu17cocoa} and PCNet-M~\cite{Zhan20deocclusion}. 

Recall is computed as the number of correctly predicted occluding orders divided by the number of ground truth occluding orders. Precision is the number of correctly predicted occluding orders divided by the total number of predicted occluding orders. F1 score is the harmonic mean of precision and recall. The equation of \emph{Recall}, \emph{Precision} and \emph{F1} score are defined as follows: 
\vspace{3mm}
\begin{equation}
    \text{Recall = } \frac{\sum_{AB}\mathbbm{1}(o'_{AB}=1 \text{ and } o_{AB}=1)}{\sum_{AB}\mathbbm{1}(o_{AB}=1)}, \\
\end{equation}
\begin{equation}
    \text{Precision = } \frac{\sum_{AB}\mathbbm{1}(o'_{AB}=1 \text{ and } o_{AB}=1)}{\sum_{AB}\mathbbm{1}(o'_{AB}=1)}, \\
\end{equation}
\begin{equation}
    \text{F1 score = } \frac{2 \times \text{Precision} \times \text{Recall}}{\text{Precision} + \text{Recall}},
\vspace{3mm}
\end{equation}
where $o$ and $o'$ denote ground truth and predicted occlusion order, and $\mathbbm{1}$ is an indicator function.

\subsection{Depth order recovery}
We evaluate depth order recovery accuracy using Weighted Human Disagreement Rate (WHDR)~\cite{bell14whdr}, which represents the percentage of weighted disagreement between ground truth $d$ and predicted depth order $d'$. The weights are proportional to the confidence of each annotation. Here, we use the inverse of \emph{count} multiplied by the minimum number of participants. WHDR evaluates \textcolor{color_depth}{\{closer, equal, farther\}} relation on each of \{distinct, overlap, or all\} categories separately; which is defined as follows:
\vspace{3mm}
\begin{equation}
\begin{split}
    \text{WHDR = } \frac{\sum_{AB} w_{AB}\cdot \mathbbm{1}(d'_{AB} \neq d_{AB})}{\sum_{AB} w_{AB}}, \\ 
    \text{ where } w_{AB} = \frac{2}{\emph{count}_{AB}}.
\end{split}
\vspace{3mm}
\end{equation}

\subsection{Disparity map prediction}
For the disparity map prediction on the KITTI dataset~\cite{Geiger2012driving}, we evaluate the performance of our InstaDepthNet and MiDaS~\cite{Ranftl2020midas} using following metrics:
\vspace{3mm}
\begin{equation}
    \text{Abs Rel = } \frac{1}{|\mathcal{G}|}\sum_{i\in \mathcal{G}}\frac{|D'(i)-D(i)|}{D(i)},\\
\end{equation}
\begin{equation}
    \text{Sq Rel = } \frac{1}{|\mathcal{G}|}\sum_{i\in \mathcal{G}}\frac{||D'(i)-D(i)||^2}{D(i)},\\
\end{equation}
\begin{equation}
    \text{RMSE log = } \sqrt{\frac{1}{|\mathcal{G}|}\sum_{i\in \mathcal{G}}||\log D'(i)-\log D(i)||^2},\\
\end{equation}
\begin{equation}
\begin{split}
    \text{Accuracy = } \text{\% of } D'(i) \text{ s.t.} \\ \max(\frac{D'(i)}{D(i)},\frac{D(i)}{D'(i)})=\delta<\tau, \\
\end{split}
\vspace{3mm}
\end{equation}
where $D$ and $D'$ denote ground truth and predicted depth maps, and $\mathcal{G}$ indicates the pixels whose ground truth values are available.

\begin{table*}[ht]
\centering
\captionsetup{width=2\columnwidth}
\resizebox{1.5\columnwidth}{!}{
\renewcommand*\arraystretch{1.2}
\begin{tabular}{l|ccc|ccc|ccc}
\toprule
 & \multicolumn{3}{c}{Occ. order}  & \multicolumn{3}{|c|}{Occ. acc. $\uparrow$}   & \multicolumn{3}{c}{Depth WHDR $\downarrow$} \\ \cline{2-10}
Methods & No & Uni & Bi      & Recall & Prec. & F1  & Distinct &Overlap& All   \\\hline\hline
PCNet-M~\cite{Zhan20deocclusion}   & $\checkmark$ & $\checkmark$  & &   62.23 &57.74	& 52.28 & - & - & -\\
OrderNet\textsuperscript{M+I}~\cite{Zhu17cocoa}  & $\checkmark$ & $\checkmark$     & &  88.68	& 62.90	& 66.23 & - & - & -\\
InstaOrderNet\textsuperscript{o}  & $\checkmark$ & $\checkmark$   &  &  89.23 &	67.08	& 69.34  & - & - & -\\
InstaDepthNet\textsuperscript{o,d}  & $\checkmark$ & $\checkmark$ &  &  79.76 & 89.39  & 78.13   & {7.09} & 23.46 & {12.79} \\
\hline  
PCNet-M~\cite{Zhan20deocclusion}  & $\checkmark$ & $\checkmark$   &  $\checkmark$ &   {59.19}    & {76.42}      & {63.02}   & - & - & -\\
OrderNet\textsuperscript{M+I} (ext.)   & $\checkmark$ & $\checkmark$   & $\checkmark$ &  84.93 &	78.21 & 77.51 & - & - & -\\ 
\rowcolor{pastelY}
InstaOrderNet\textsuperscript{o} & $\checkmark$ & $\checkmark$ & $\checkmark$ & 89.39 & 79.83 & 80.65 & - & - & -\\ 
\rowcolor{pastelY}
InstaDepthNet\textsuperscript{o,d} &   $\checkmark$   & $\checkmark$ & $\checkmark$ &      \textbf{84.89} & \textbf{91.34} & \textbf{85.01} &  \textbf{7.00} & \textbf{23.29} & \textbf{12.72}  \\ \bottomrule
\end{tabular}}
\caption{Ablation study of utilizing bidirectional occlusion orders for occlusion and depth order prediction tasks.}
\label{table_bidirec}      
\end{table*}

\begin{table*}[ht]
\begin{center}
\captionsetup{width=2\columnwidth}
\resizebox{1.7\columnwidth}{!}{
\renewcommand*\arraystretch{1.2}
\begin{tabular}{ccc||ccc|ccc}
\toprule
\multicolumn{3}{c||}{Loss weights} & \multicolumn{3}{c|}{InstaOrder} & \multicolumn{3}{c}{DIW}                           \\\hline
$\mathcal{L}_{do}$      & $\mathcal{L}_{disp}$      & $\mathcal{L}_{s}$       & WHDR Distinct $\downarrow$&WHDR Overlap $\downarrow$ & WHDR All $\downarrow$    & Correct $\uparrow$         & Wrong $\downarrow$          & WHDR $\downarrow$          \\\hline    \hline
1        & 0          & 0.1      & 7.24                                 & 23.99                        & 13.17                         & 65,270          & 9,171          & 12.32          \\
1        & 1          & 0        & \textbf{7.14} & 23.64 & 13.00 & 65,277          & 9,164          & 12.30 \\
\rowcolor{pastelY} 1        & 1          & 0.1      & 7.25 & \textbf{23.34}               & \textbf{12.94}               & \textbf{65,317} & \textbf{9,124} & \textbf{12.26} \\ \bottomrule
\end{tabular}}
\end{center}
\vspace{-4mm} 
\caption{Ablation study on losses applied to InstaDepthNet\textsuperscript{d}.}
\label{table_loss}
\end{table*}

\section{Additional results}
\subsection{Bidirectional occlusion order}
We conduct experiments with the \textsc{InstaOrder} dataset to verify the effect of bidirectional occlusion orders. We compare the accuracy with and without using the bidirectional occlusion orders for both training and testing (Table~\ref{table_bidirec}). Intuitively, classifying smaller occlusion order categories \textcolor{color_occ}{(no occlusion, A$\rightarrow$B, B$\rightarrow$A)} seems more manageable, but methods not using bidirectional order reported lower scores than those using bidirectional order. We speculate that bidirectional order helps to distinguish ambiguous ordering cases.

\subsection{Loss functions}
We conduct an ablation study on loss functions to validate the effectiveness of our proposed instance-wise disparity loss. We train InstaDepthNet\textsuperscript{d} with varying losses using \textsc{InstaOrder} training set. Then we report WHDR using \textsc{InstaOrder} validation set and DIW test set. $L_{do}$ is depth order loss, $L_{disp}$ is the proposed instance-wise disparity loss, and $L_s$ is edge-aware smoothness loss (Sec 4.2 in the main paper). Experimental result (Table~\ref{table_loss}) shows that accuracy degraded without $L_{disp}$ or $L_s$. Especially, the absence of $L_{disp}$ degraded the accuracy by a large margin, which demonstrates the usefulness of the proposed instance-wise disparity loss.

\begin{figure*}[hbt]
\centering
\begin{subfigure}{1.2\columnwidth}
  \centering
  \includegraphics[height=3in]{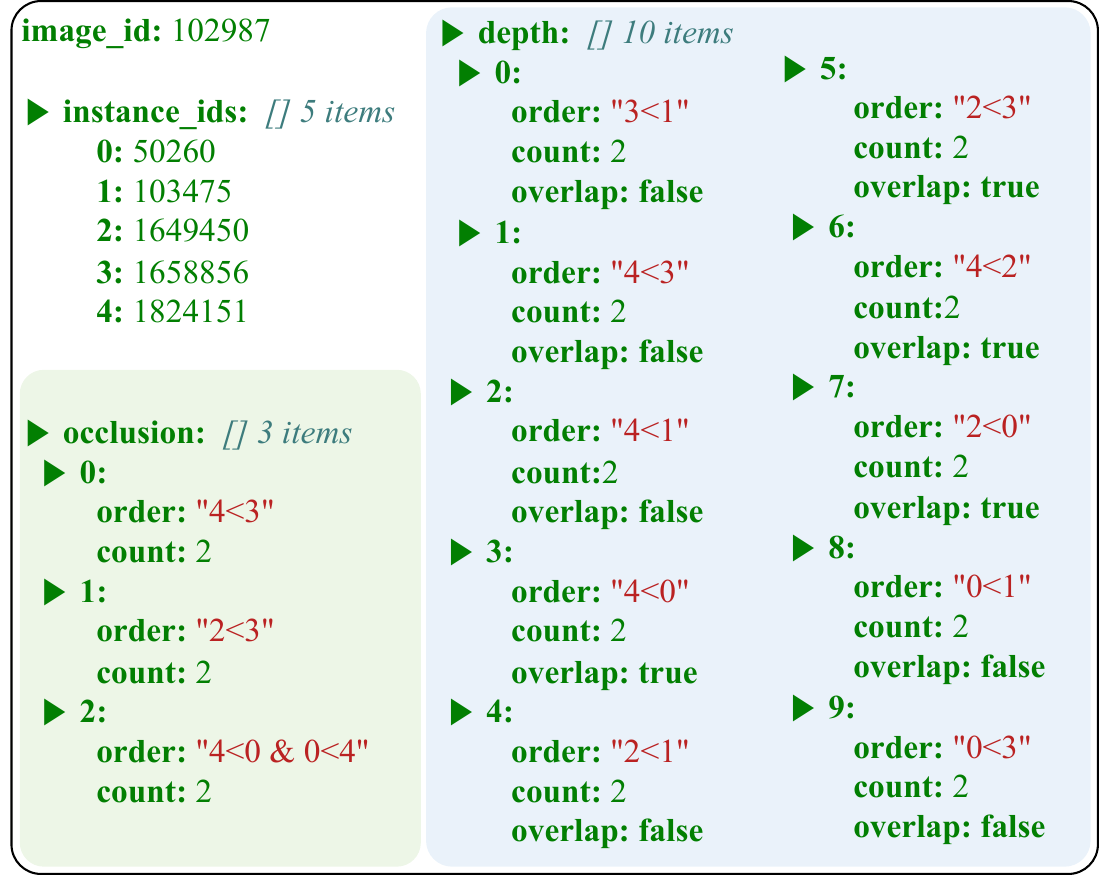}
  \caption{\texttt{json} file provided by \textsc{InstaOrder} dataset}
\end{subfigure}
\begin{subfigure}{0.8\columnwidth}
  \centering
  \includegraphics[height=3in]{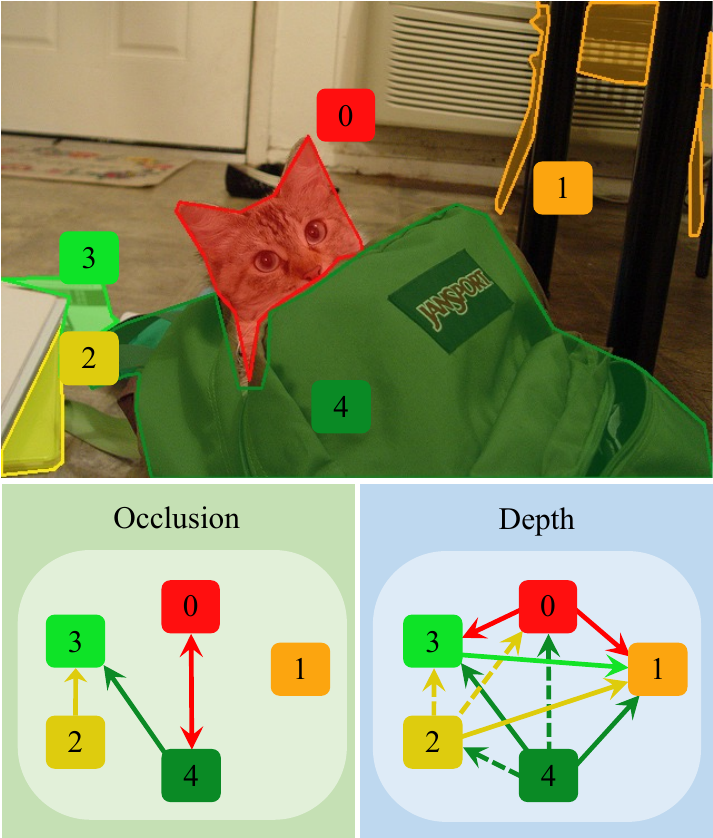}
  \caption{Instance-wise occlusion and depth orders}
\end{subfigure}
\vspace{-2mm}
\caption{Overview of the proposed \textsc{InstaOrder} dataset.}
\vspace{-2mm}
\label{fig:instaorder_json}
\end{figure*}

\section{\textsc{InstaOrder} Information}
\subsection{License}
We constructed the \textsc{InstaOrder} dataset utilizing COCO 2017~\cite{Lin14coco} images and instance masks. COCO 2017 annotations are licensed under a CC BY 4.0 license. Image source of COCO 2017 is Flickr, and the copyrights follow Flickr's terms of use\footnote{\url{https://www.flickr.com/creativecommons/}}. Similarly, our annotations in \textsc{InstaOrder} are licensed under a CC BY 4.0 license. 

\subsection{Guideline}
We provide a guideline to annotators: we ask them to annotate only semantically meaningful instances and to consider the entire structure of instances. Some instance pairs are unclear to annotate occlusion and depth order. For example, (\romannumeral 1) a collage image (multiple photos appear in one photo) and (\romannumeral 2) objects shown on television, magazine, or mirror. Images of case (\romannumeral 1) are discarded, and instances in case (\romannumeral 2) are annotated as equally distant without occlusion. To eliminate the bias from the image sequence, we provide randomly shuffled images for each annotator.

\subsection{Wages}
We annotated \textsc{InstaOrder} by crowd-sourcing, and the total amount of money given to workers is \$35,000. The workers are paid based on the number of annotations. Before the crowd-sourcing begins, we monitor unprofessional workers and measure the average time taken for one annotation to set the proper reward. We set different rewards for occlusion and depth order annotation based on the time. 

After crowd-sourcing finishes, we check the actual annotation times by crowd workers. On average, it took 2.68 seconds for a single occlusion order annotation and 5.05 seconds for a single depth order annotation. With this speed, the hourly rewards we give are \$6 for the occlusion order task and \$4.5 for the depth order task. We also provide \$45 for each of the top 50 depth order annotators to promote the task. This reward design motivated crowd workers, and our task was popular on the crowd-sourcing platform. As a result, a total of 3,659 workers participated in the task, and the annotation job just took a month.

\subsection{Data example}
As noted, \textsc{InstaOrder} is annotated upon COCO 2017~\cite{Lin14coco} dataset, and therefore only the \texttt{json} file is provided (Figure~\ref{fig:instaorder_json}, a). For an image (image\_id), five instances (instance\_ids) with class labels are from the COCO dataset. With this information, we denote the occlusion order as "occluder\_id $<$ occludee\_id", and for bidirectional order "A$<$B \& B$<$A" notation is used. Similarly, depth order is denoted as "closer\_id $<$ farther\_id" and for equal depth "A=B" notation is used. Besides the orders, we also provide the metadata such as count and overlap. As a result, we can generate occlusion and depth graphs (Figure~\ref{fig:instaorder_json}, b). 
\section{Implementation Details}
\subsection{Training details}\label{implementation_details}
We train InstaOrderNet with SGD optimizer~\cite{Bottou10sgd} for 58K iterations. The initial learning rate set to 0.001 is decayed by 0.1 after 32K and 48K iterations. InstaOrderNet use ResNet-50~\cite{He16resnet} initialized with Xavier init~\cite{GlorotB10xavier}. We use a batch size of 128 distributed over four Nvidia TITAN RTX GPUs.

InstaDepthNet consists of two heads: the order prediction head and the depth map prediction head. The order prediction head uses ResNet-50~\cite{He16resnet} and the depth map prediction head uses MiDaS-v2~\cite{Ranftl2020midas}. Similar to InstaOrderNet, ResNet-50~\cite{He16resnet} is utilized with Xavier init~\cite{GlorotB10xavier}. For MiDaS-v2~\cite{Ranftl2020midas}, we adopt pre-trained weights\footnote{\url{https://github.com/intel-isl/MiDaS}} provided by the authors. We train InstaDepthNet with a batch size of 48 using four Nvidia V100 GPUs. The initial learning rate is set to 0.0001.

\subsection{InstaOrderNet architecture}
\begin{figure}[htb]
  \centering
  \includegraphics[width=1\columnwidth]{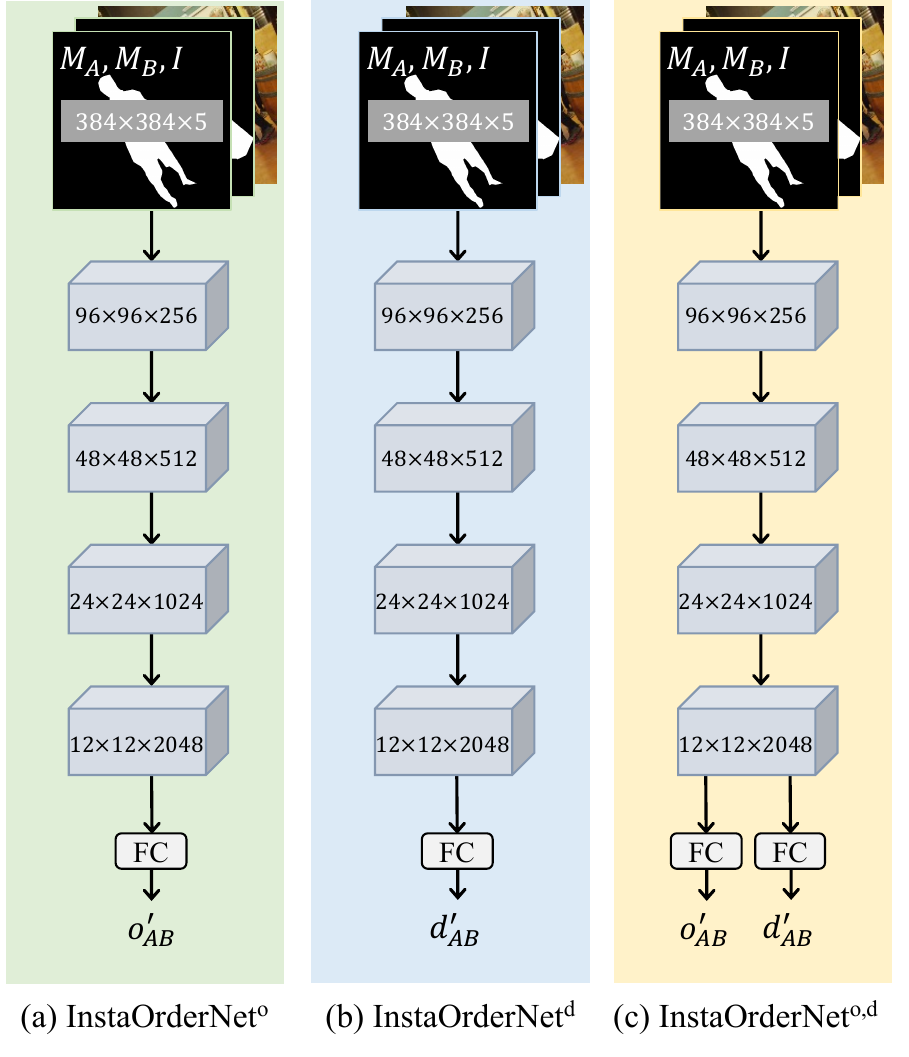}
    \vspace{-6mm}
    \caption{InstaOrderNet model architecture.}
    \vspace{-6mm}
    \label{fig:InstaOrderNet_arch}
\end{figure}

InstaOrderNet (Figure~\ref{fig:InstaOrderNet_arch}) takes pairwise instance masks ($M_A, M_B$) and an image ($I$) as input and outputs their instance-wise orders. InstaOrderNet uses ResNet-50~\cite{He16resnet} as noted. Here we denote the feature size by [\texttt{height}, \texttt{width}, \texttt{channel}].

\begin{figure*}[ht]
    \centering
    \includegraphics[width=1.6\columnwidth]{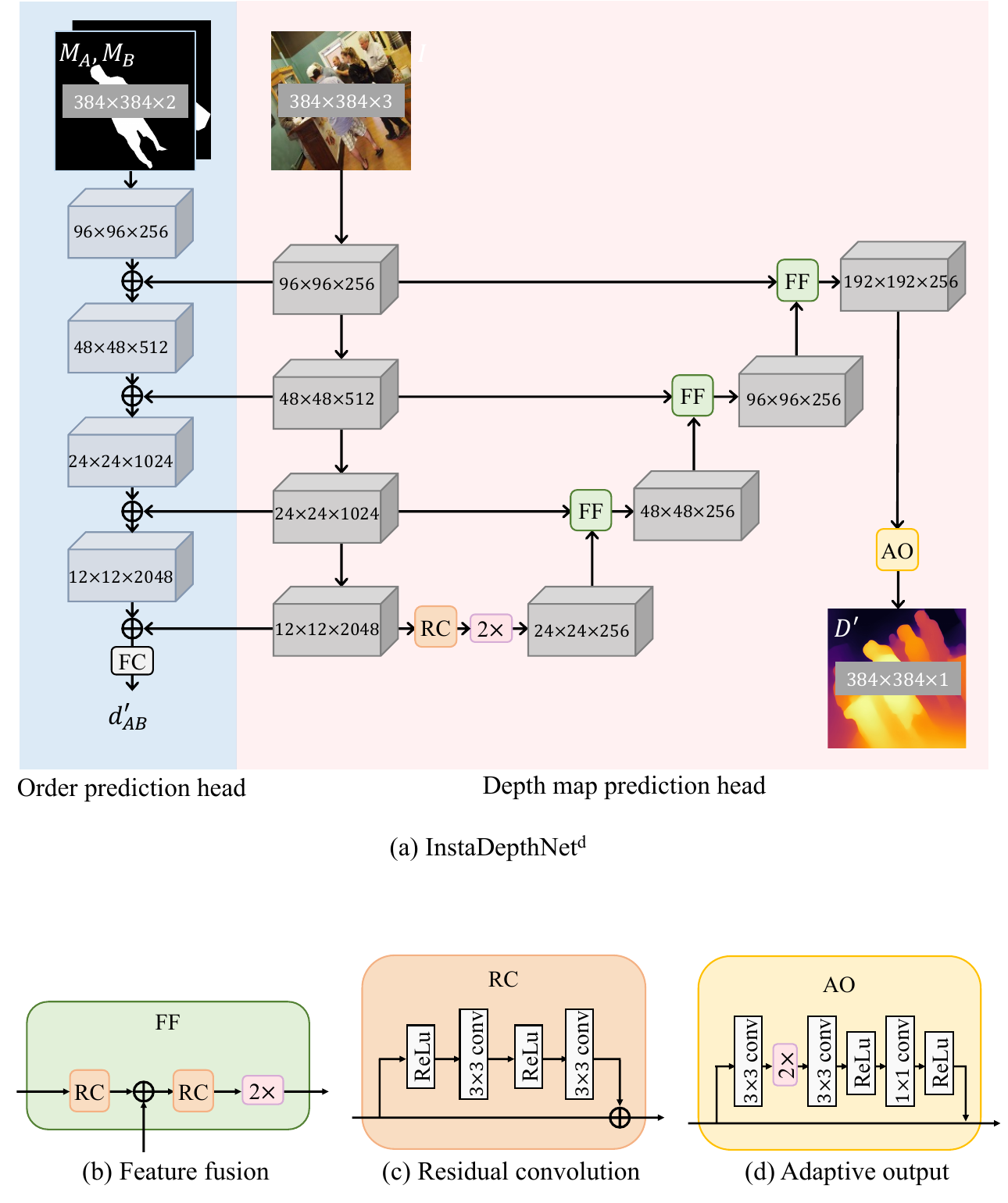}
    \caption{InstaDepthNet\textsuperscript{d} model architecture.}
    \label{fig:InstaDepthNet_arch}
\end{figure*}

\subsection{InstaDepthNet architecture}
InstaDepthNet\textsuperscript{d} (Figure~\ref{fig:InstaDepthNet_arch}) takes pairwise instance masks ($M_A, M_B$) and an image ($I$) as input and outputs their depth order($d^{'}_{AB}$) along with a disparity map ($D^{'}$). The depth order prediction head uses ResNet-50~\cite{He16resnet}, and the disparity map prediction head is MiDaS~\cite{Ranftl2020midas}. Please refer to the paper by Xian~\etal~\cite{Xian18redweb} for a detailed explanation of MiDaS architecture.

InstaDepthNet\textsuperscript{d} architecture is modular because the order prediction head can be used optionally depending on the requirements at test time. Specifically, InstaDepthNet\textsuperscript{d} can produce a dense disparity map even when instance masks are unavailable, such as DIW~\cite{Chen16diw} dataset. 

\subsection{Input resolution.} 
For the networks that produce depth order (InstaOrder\textsuperscript{d}, InstaOrder\textsuperscript{o,d}, InstaDepthNet\textsuperscript{d} and InstaDepthNet\textsuperscript{o,d}), we set image resolution as $384\times384$ by following the MiDaS~\cite{Ranftl2020midas}. On the other hand, for the network that does not produce depth order (InstaOrderNet\textsuperscript{o}), we set the input size as described in PCNet-M~\cite{Zhan20deocclusion}. Inputs of InstaOrderNet\textsuperscript{o} are patches that are adaptively cropped to contain objects at the center, then resized to $256\times256$ at the train and test time.  

\section{License of Other Assets}
COCO 2017~\cite{Lin14coco} annotations are licensed under a CC BY 4.0 license. Image source of COCO 2017 is Flickr, and the copyrights follow Flickr's terms of use\footnote{\url{https://www.flickr.com/creativecommons/}}. We conducted experiments using the dataset COCOA~\cite{Zhu17cocoa}, KINS~\cite{Qi19kins}, and DIW~\cite{Chen16diw}. To our best knowledge, COCOA and KINS are publicly released as written in the papers~\cite{Zhu17cocoa, Qi19kins}. However, we could not find the appropriate license for the DIW~\cite{Chen16diw} dataset.

We utilized a pre-trained model of MiDaS-v2~\cite{Ranftl2020midas}\footnote{\url{https://github.com/intel-isl/MiDaS}} that follows MIT license, and PCNet-M~\cite{Zhan20deocclusion}\footnote{\url{https://github.com/XiaohangZhan/deocclusion/}} that follows Apache License 2.0.

\newpage
{\small
\bibliographystyle{ieee_fullname}
\bibliography{egbib}
}

\end{document}